\documentclass[10pt,twocolumn,letterpaper]{article}

\usepackage{cvpr}
\usepackage{times}
\usepackage{epsfig}
\usepackage{graphicx}
\usepackage{amsmath}
\usepackage{amssymb}

\usepackage{array}
\usepackage{color}
\usepackage{multirow}
\usepackage{hhline}
\usepackage{subfigure}
\usepackage{algorithm}
\usepackage{algorithmic}
\usepackage{amsmath}
\usepackage{amssymb}
\usepackage{amsfonts}
\usepackage{amsthm}
\usepackage{subfigure}
\usepackage{listings}

\usepackage[pagebackref=true,breaklinks=true,letterpaper=true,colorlinks,bookmarks=false]{hyperref}

\cvprfinalcopy 

\def\cvprPaperID{1738} 

\ifcvprfinal\fi
\begin{document}
\graphicspath{ {./figures/} }
 
\title{Regularizing RNNs for Caption Generation by\\ Reconstructing The Past with The Present}

\author{Xinpeng Chen$^{\dagger}$\thanks{This work was done while Xinpeng Chen was a Research Intern with Tencent AI Lab.} \qquad Lin Ma$^\ddagger$\renewcommand{\thefootnote}{\fnsymbol{footnote}}\footnotemark[4] \qquad Wenhao Jiang$^\ddagger$\renewcommand{\thefootnote}{\fnsymbol{footnote}}\footnotemark[4] \qquad Jian Yao$^{\dagger}$ \qquad Wei Liu$^\ddagger$ \\
$^\dagger$Wuhan University \qquad  $^\ddagger$Tencent AI Lab\\
{\tt\small \{jschenxinpeng, forest.linma, cswhjiang\}@gmail.com} \\
{\tt\small jian.yao@whu.edu.cn} \qquad
{\tt\small wliu@ee.columbia.edu}}


\maketitle
\thispagestyle{empty}
\renewcommand{\thefootnote}{\fnsymbol{footnote}}
\footnotetext[4]{Corresponding authors.}
%
\begin{abstract}
Recently, caption generation with an encoder-decoder framework has been extensively studied and applied in different domains, such as image captioning, code captioning, and so on. In this paper, we propose a novel architecture, namely Auto-Reconstructor Network (ARNet), which, coupling with the conventional encoder-decoder framework, works in an end-to-end fashion to generate captions. ARNet aims at reconstructing the previous hidden state with the present one, besides behaving as the input-dependent transition operator. Therefore, ARNet encourages the current hidden state to embed more information from the previous one, which can help regularize the transition dynamics of recurrent neural networks (RNNs). Extensive experimental results show that our proposed ARNet boosts the performance over the existing encoder-decoder models on both image captioning and source code captioning tasks. Additionally, ARNet remarkably reduces the discrepancy between training and inference processes for caption generation. Furthermore, the performance on permuted sequential MNIST demonstrates that ARNet can effectively regularize RNN, especially on modeling long-term dependencies. Our code is available at: \url{https://github.com/chenxinpeng/ARNet}.
\end{abstract}

\section{Introduction}

\begin{figure}[!t]
    \centering
    \includegraphics[width=\linewidth]{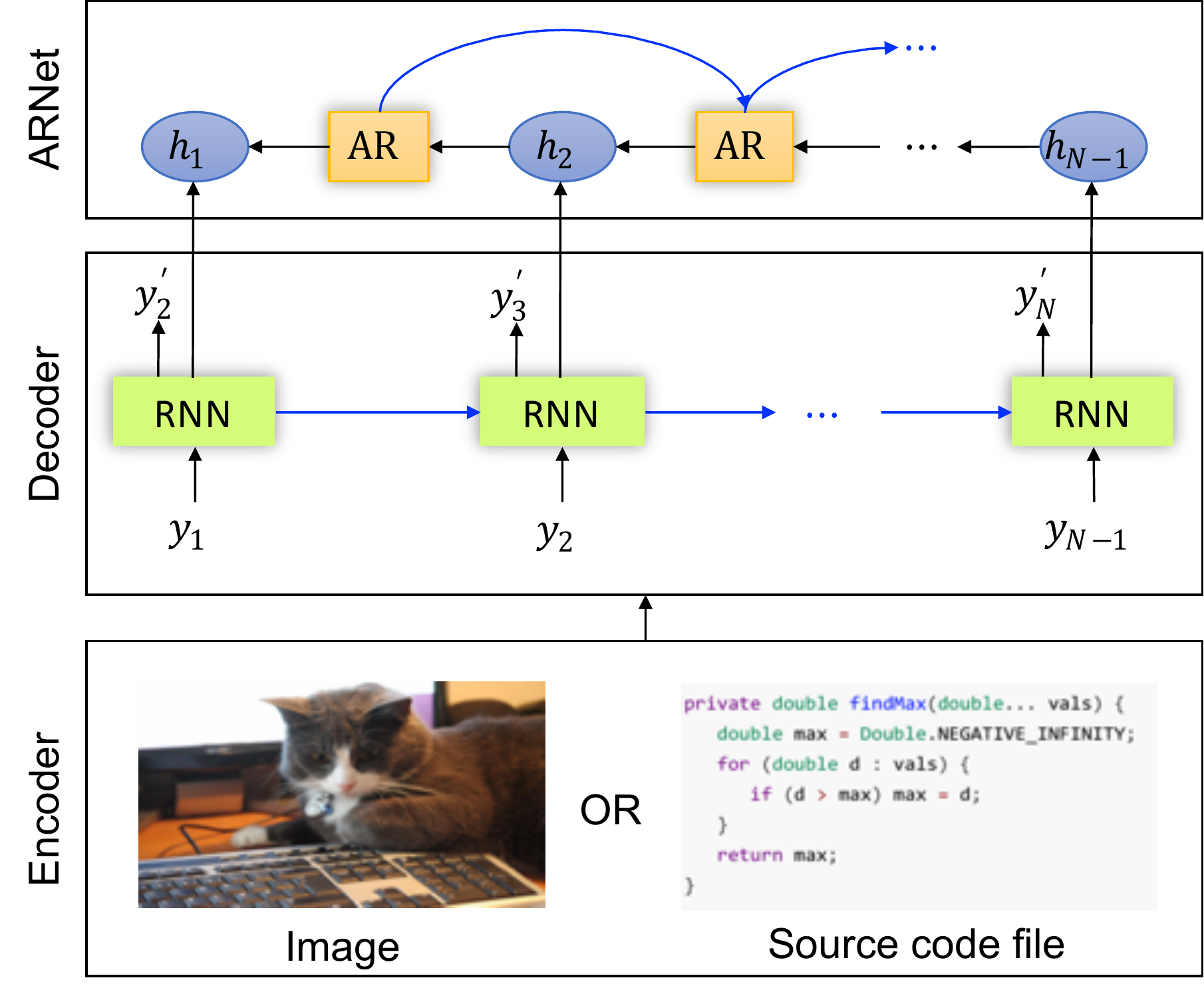}
    \caption{An overview of our proposed ARNet coupling with the conventional encoder-decoder framework. The encoder takes an image or a source code file as input and generates its semantic embedding, based on which the decoder, usually one RNN, can thus generate the semantically-correlated captions. Other than an input-dependent transition operator used in the decoder, the proposed ARNet connects the neighboring hidden states together by reconstructing the past hidden state with the present one. The blue arrows indicate the state transitions in RNN.}
    \label{fig1}
\end{figure}

Caption generation~\cite{ReviewnetNIPS2016,chen2016sca} is a fundamental research problem, which has received increasing attention in both computer vision and natural language processing communities. The task is to predict a syntactically and semantically correct target sequence consisting of consecutive words based on the provided source information. For example, an image captioning task aims to generate an appropriate sentence to describe the image content~\cite{Vinyals_2015,KarpathyPAMI2017}, while a code captioning task targets at providing a sentence to summarize the conceptual idea behind the given source code file~\cite{Iyer_2016,ReviewnetNIPS2016}. Caption generation is a very challenging task. First, the semantic meaning of the given source needs to be well learned and captured, especially for different modalities, such as image and source code. Second, the target sentence generating process needs to not only maintain the syntactical correctness but also ensure the semantic correlations with the source information, which thus requires complicated interactions between them.

Recent work on caption generation, such as image captioning~\cite{Vinyals_2015}, counts on an encoder-decoder framework to generate the corresponding sentence for one given image. As illustrated in Fig.~\ref{fig1}, the encoder takes one image or source code file as input and generates its corresponding semantic embedding. Due to the different behaviors and characteristics of the source, different neural network architectures are used as the encoder, \emph{e.g.}, convolutional neural networks (CNNs) for images and recurrent neural networks (RNNs) for sequential data such as source code and natural language. With the semantic embedding, the decoder employs another RNN to generate the target sentence to reflect the content of the image or summarize the conceptual idea of the source code. Moreover, in order to encourage the decoder to focus on the crucial information for generating captions, attention mechanisms were proposed for image captioning~\cite{XuICML2015} and text abstractive summarization~\cite{Iyer_2016,Chopra_2016}. At each time step, the attention strategy measures the relevance of the encoder's hidden states given all the previously generated words in the target sentence. However, the attention mechanism proceeds in a sequential manner, which lacks global modeling capacities. In order to address this drawback, a review network~\cite{ReviewnetNIPS2016} was proposed with review steps lying between the encoder and the decoder. As such, a more compact, abstractive, and global annotation vectors are generated, which have been demonstrated to further benefit the sentence generation process.

Even though the encoder-decoder architecture and its variants have achieved remarkable performance improvements on caption generation tasks, two problems still remain. First, the decoder relies on an input-dependent transition operator to generate captions. Specifically, the word ${y^{\prime}_{t+1}}$ is conditioned on the hidden state $h_{t}$ at time step $t$ independently, which has not fully exploited the latent relationship with its previous one $h_{t-1}$.
Second, the discrepancy, also named as exposure bias, in RNN between training and inference still exists~\cite{ProfessorForcing,ScheduledSampling}. During the training phase, we take the ground-truth word $y_t$ as input of the RNN unit to predict the next word $y^{\prime}_{t+1}$ and force it to stay close to $y_{t+1}$. However, the ground-truth word $y_{t}$ is unavailable during the inference phase. The RNN unit depends on the generated word $y^{\prime}_{t}$  by the model from the previous time step  for  $y^{\prime}_{t+1}$  prediction.

In order to handle the aforementioned problems, in this paper, we introduce an Auto-Reconstructor Network (ARNet) coupling with the conventional encoder-decoder framework for caption generation, as illustrated in Fig.~\ref{fig1}. Our proposed ARNet connects two neighbouring hidden states by reconstructing the past hidden state  with present one. As such, ARNet encourages the current hidden state to embed more information from the previous one. The transition dynamics of the RNN in the decoder are thus regularized. Our main contributions lie in three-fold:
\begin{itemize}
  \item We propose a novel architecture that introduces an ARNet coupling with the encoder-decoder framework, which strengthens the connection between neighboring hidden states by reconstructing the past with the present.
  \item ARNet can help regularize the transition dynamics of the RNN, therefore mitigating its discrepancy for sequence prediction.
  \item ARNet coupling with the encoder-decoder framework and its variants achieve performance improvements on both image captioning and source code captioning tasks. Moreover, ARNet, conducting regularization on RNN, can effectively model long term dependencies.
\end{itemize}

\section{Related Work}
\subsection{Encoder-Decoder Framework}
The encoder-decoder framework for caption generation is inspired by its successful application to machine translation~\cite{Cho_2014}, where RNNs were used for both encoding and decoding. Generally, in an encoder-decoder framework, the encoder encodes the input into an informative vector and the decoder translates the vector into a corresponding sequence. Either image captioning or code captioning can be seen as a task of translation. And the encoder-decoder framework has achieved a great success on these tasks~\cite{Vinyals_2015,ReviewnetNIPS2016,jiang_ltg}. To allow the RNN unit to determine which sub-part of input data is more important for each time step, the attention mechanism was introduced in the encoder-decoder framework and remarkably improved the performance \cite{XuICML2015}. Thereafter, many extensions of attention mechanism have been proposed \cite{You_2016,Mun_Text-Guided_2017} to push the limits of this framework for caption generation tasks.

\subsection{Exposure Bias and Regularization for RNN}
An inevitable problem for sequence generation tasks is exposure bias when the network is trained with the \emph{teacher forcing} technique~\cite{Williams_1989}. Scheduled sampling~\cite{ScheduledSampling} introduces a sampling mechanism to imitate the sequence prediction process during the training phase. While scheduled sampling has achieved good performance on the image captioning task, Huszar~\cite{Huszar2015HowT} demonstrated that this training technique is not a consistent estimation strategy. Furthermore, the professor forcing~\cite{ProfessorForcing} used generative adversarial networks~\cite{GoodFellowGAN2014} to encourage the distributions of recurrent hidden states of training and inference phase to match with each other. Recently, Krueger $et~al.$ proposed zoneout~\cite{ZoneoutICLR2017} to regularize RNN. The values of the hidden states and memory cells of the RNN either maintain their previous values or are updated as usual. Therefore, stochastic identity connections between subsequent time steps were introduced in zoneout. Note that the information of the previous hidden state randomly enters the current time step in zoneout. In contrast, our model encourages the current hidden state to absorb information from a previous time step by forcing the current hidden state to reconstruct the previous one.

\section{Background}
ARNet is proposed to couple with the encoder-decoder framework to improve the performance of caption generation tasks. In this section, we briefly review the encoder-decoder framework.

\subsection{Encoder}
In the encoder-decoder framework, the encoder is used to generate the semantic representation of input data. In order to make a full understanding of the input data, the encoder generates not only the global information in the form of one distributed vector $g$ but also the local information represented by a set of vectors $\mathbf{s} =\{s_1,s_2,\dots,s_n\}$, which will be further used as the input of the decoder.

Due to different behaviors and characteristics of the source input, different types of encoders have been used for different caption generation tasks. For image captioning, recently developed CNNs, such as Inception-X~\cite{InceptionV1,InceptionV2,InceptionV3,InceptionV4} and ResNet~\cite{ResNet}, are usually utilized to generate global and local representations of images. In this paper, we employ Inception-V4 to encode one given image $\mathbf{I}$, with the last fully-connected layer being the global representation  $g$ and the outputs of the last convolutional layer composing the local information vectors $\mathbf{s}$, respectively.

For the task of source code captioning \cite{ReviewnetNIPS2016}, RNNs are more naturally suited for modeling the source code sequence. Given one input source code token sequence $\mathbf{I} = \{ i_1, ..., i_{T} \}$, at time step $j$ we feed $i_j$ into the RNN unit and obtain the hidden state $h_{j}$. The hidden state of the last time step $h_T$ encodes the information of the whole sequence, which is regarded as containing the sequence global information. And the hidden states generated during the encoding process contain the subsequence information, which are composed as local information vectors. In order to well capture long term dependencies, long short-term memory (LSTM)~\cite{Schmidhuber_1997} and gated recurrent unit (GRU)~\cite{Cho_2014} with specifically designed gating mechanisms were proposed. In this paper, LSTM is employed as the encoder for handling input sequence data.

LSTM unit acts as a transition operator transferring the previous hidden state $h_{t-1}$ to the current hidden state $h_t$ with the input $x_t$ at time $t$:
\begin{align}
  h_t = \textrm{LSTM}(x_t, h_{t-1}).
\end{align}
In this paper, we use the same definitions as \cite{Zaremba_2014}. Then the LSTM transition process can be formulated as follows:
\begin{equation}
  \small
  \label{lstm_linear}
  \begin{split}
    \begin{pmatrix} i_t \\ f_t \\ o_t \\ g_t \end{pmatrix} &=
    \begin{pmatrix} \sigma \\ \sigma \\ \sigma \\ \tanh \end{pmatrix}
    \mathbf{T}
    \begin{pmatrix} x_t \\ h_{t-1} \end{pmatrix},\\
    c_t &= f_t \odot c_{t-1} + i_t \odot g_t, \\
    h_t &= o_t \odot \tanh(c_t),
  \end{split}
\end{equation}

\noindent where $i_t$, $f_t$, $o_t$, $c_t$, $h_t$, and $\sigma$ are input gate, forget gate, output gate, memory cell, hidden state, and sigmoid function, respectively. $\mathbf{T}$ is a linear transformation matrix. $\odot$ represents an element-wise product operator.

\subsection{Decoder}
Based on the global information vector $g$ and local information vectors $\mathbf{s}$ generated by the encoder, the aim of the decoder is to generate a natural sentence $\mathcal{C}$ consisting of $N$ words $(y_1,y_2,\cdots, y_N)$, which not only expresses content information of the input source, \emph{e.g.}, image or source code, but also should be naturally coherent. To further exploit the contributions of the local information vectors and improve the performance, the attention mechanism~\cite{Bahdanau_2014,XuICML2015} was proposed. Therefore, the attentive LSTM can be further reformulated as:
\begin{equation}
  \begin{split}
	  \begin{pmatrix} i_t \\ f_t \\ o_t \\ g_t \end{pmatrix} &=
  \begin{pmatrix} \sigma \\ \sigma \\ \sigma \\ \tanh \end{pmatrix}
  \mathbf{T}
  \begin{pmatrix} x_t \\ h_{t-1} \\ z_{t} \end{pmatrix},
  \end{split}
\end{equation}
where $z_{t}$ denotes the context vector, yielded by the attention mechanism. Given the local information vectors $\mathbf{s}$ generated from the encoder, $z_{t}$ is computed by:
\begin{align}
  z_{t} = f_{att}(\mathbf{s}, h_{t-1}) = \sum_{i=1}^{|\mathbf{s}|}{\frac{\exp\left(\alpha(s_i, h_{t-1})\right)}{\sum\nolimits_{j=1}^{|\mathbf{s}|}{\exp\left(\alpha(s_{j}, h_{t-1})\right)}} s_{i}}.
\end{align}
$\alpha(s_i, h_{t-1})$ measures the similarity between $s_i$ and $h_{t-1}$, which is usually realized by a multilayer perceptron. LSTMs with or without the attention mechanism can both be used as the decoder. In this paper, in order to demonstrate the effectiveness of our proposed ARNet, we experiment on two LSTMs, attentive LSTM and LSTM without attention.

\section{The Proposed ARNet}
\subsection{Architecture}
\label{sec:architecture}

\begin{figure}[!t]
    \centering
    \includegraphics[width=\linewidth]{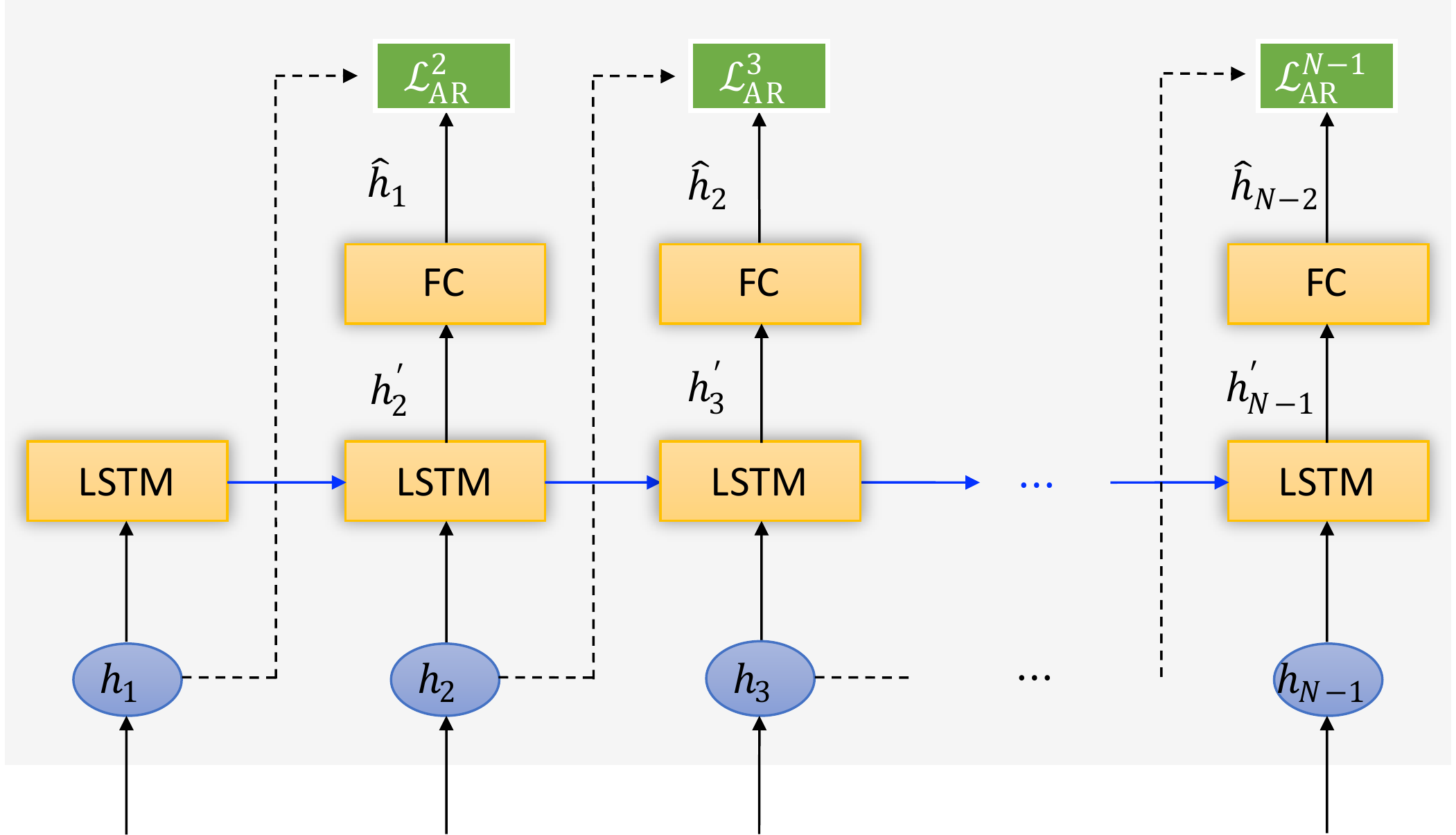}
    \caption{The framework of our proposed ARNet. At each time step of the decoder, ARNet takes the present hidden state $h_t$ as the input to reconstruct its previous hidden state $h_{t-1}$. $\mathcal{L}_{\text{AR}}^{t}$ is defined to match the reconstructed output $\hat{h}_{t-1}$ and the previous hidden state $h_{t-1}$.}
    \label{fig2}
\end{figure}

As shown in Fig.~\ref{fig1}, the proposed ARNet couples with the encoder-decoder framework for caption generation. Concretely, our proposed ARNet is realized by another LSTM, taking the hidden states sequence yielded in the decoder as inputs. The architecture of ARNet is illustrated in Fig.~\ref{fig2}, from which we can see that ARNet aims at exploiting the relationships between neighboring hidden states.

LSTM unit is leveraged to reconstruct the past hidden state $h_{t-1}$ with the present one $h_t$, which can be formulated as:

\begin{equation}
  \begin{split}
  \begin{pmatrix} i_t^{\prime} \\ f_t^{\prime} \\ o_t^{\prime} \\ g_t^{\prime} \end{pmatrix} &=
  \begin{pmatrix} \sigma \\ \sigma \\ \sigma \\ \tanh \end{pmatrix}
  \mathbf{T}
  \begin{pmatrix} h_{t} \\ h_{t-1}^{\prime} \end{pmatrix}, \\
  c_t^{\prime} &= f_t^{\prime} \odot c_{t-1}^{\prime} + i_t^{\prime} \odot g_t^{\prime}, \\
  h_t^{\prime} &= o_t^{\prime} \odot \tanh(c_t^{\prime}),
  \end{split}
\end{equation}

\noindent where $i_t^{\prime}$, $f_t^{\prime}$, $o_t^{\prime}$, $c_t^{\prime}$ and $h_t^{\prime}$ are the input gate, forget gate, output gate, memory cell and hidden state of the LSTM unit, respectively. In order to further match the previous hidden state $h_{t-1}$, one fully-connected layer is employed to map the generated $h_t^\prime$ into the common space with $h_{t-1}$:

\begin{align}
  \hat{h}_{t-1} = \mathbf{w}_{\mathbf{fc}}h^\prime_t + \mathbf{b}_{\mathbf{fc}},
\end{align}

\noindent where $\mathbf{w}_{\mathbf{fc}}$ and $\mathbf{b}_{\mathbf{fc}}$ are the weight matrix and bias vector, respectively. $\hat{h}_{t-1}$ is the reconstructed previous hidden state. Afterwards, we define a reconstruction error in terms of Euclidean distance between $h_{t-1}$ and $\hat{h}_{t-1}$:

\begin{align}
  \mathcal{L}_{\text{AR}}^{t} = \parallel h_{t-1} - \hat{h}_{t-1}\parallel ^2_2,
\end{align}

\noindent where $\mathcal{L}_{\text{AR}}^{t}$ measures the reconstruction error of the ARNet at time step $t$. Through minimizing the defined reconstruction error, we encourage the current hidden state $h_t$ to embed more information from the previous one $h_{t-1}$. 

Such a reconstruction strategy in our proposed ARNet, behaving similarly to the zoneout regularizer~\cite{ZoneoutICLR2017}, regularizes the LSTM during the caption generation process. Zoneout regularizes RNNs by randomly preserving hidden activations, which stochastically forces some parts of hidden unit and memory cell to maintain their previous values at each time step. With such a process, gradient and state information are more steadily propagated through time~\cite{ZoneoutICLR2017}. However, zoneout can be regarded as one ``hard'' strategy, which stochastically makes a binary choice between previous and current hidden states. On the contrary, the reconstruction strategy of our ARNet presents to be one ``soft'' scheme, which learns to adaptively embed the information of the previous hidden state into the current one. Therefore, the ARNet relies on LSTM to adaptively fuses both the previous and current hidden states together, rather than randomly chooses the previous or current one. 

Moreover, with the ARNet reconstructing $h_{t-1}$ from $h_{t}$, we encourage the backward information to flow through the network, as shown in Fig.~\ref{fig1}. The correlations between $h_t$ and $h_{t-1}$ are further exploited and enhanced. In doing so, the transition dynamics through time on the LSTM is regularized. Furthermore, since the ARNet couples with the encoder-decoder framework, the exposure bias problem in sequence generation can be alleviated, which will be demonstrated and discussed in the following experimental section.

\subsection{Training Procedure}

The training procedure of our model consists of two stages. First, we freeze the parameters of the ARNet and pre-train the encoder-decoder architecture, which is usually trained by the negative log-likelihood:
\begin{align}
  \label{obj_lm}
  \mathcal{L}_{\text{NLL}} = -\log\ p(\mathcal{C}|\mathcal{I}) = -\sum_{t=2}^{N}  \log p(y_{t}|y_{t-1}),	
\end{align}
where $\mathcal{I}$ is the input source, particularly the image or source code,  $\mathcal{C}$ denotes the generated caption given $\mathcal{I}$, $p(y_{t}|y_{t-1}) = \textrm{Softmax}(\mathbf{w}h_t)$, with $\mathbf{w}$ being the linear transformation matrix, and $h_t = \textrm{LSTM}(\mathbf{E}\mathbf{y}_{t}, h_{t-1})$.  $y_1$ is the sign for the start of a sentence. And $\mathbf{E}\mathbf{y}_{t}$ denotes the distributed representation of the word $\mathbf{y}_{t}$, where $\mathbf{y}_{t}$ is the one-hot representation for the word $y_{t}$ and $\mathbf{E}$ is the word embedding matrix. After the encoder-decoder architecture converges, the whole network is fine-tuned using the following objective function: 
\begin{align}
  \label{obj_whole}
  \mathcal{L} =  \mathcal{L}_{\text{NLL}} + \lambda \sum \mathcal{L}_{\text{AR}}^t.
\end{align}
Here, $\lambda$ is a trade-off parameter to balance the contributions from the ARNet and the encoder-decoder architecture. 


\begin{table*}[t]
  \caption{Single model performance of a variety of models on Karpathy's splits of the MSCOCO dataset. The highest entry for each evaluation metric is highlighted in boldface. }\label{table_1}
  \begin{center}
    \footnotesize
    \begin{tabular}{|l|c|c|c|c|c|c|c|c|}
      \hline
      Model Name & BLEU-1 & BLEU-2 & BLEU-3 & BLEU-4 & METEOR & ROUGE-L & CIDEr & SPICE \\
      \hline
      \hline
      NIC~\cite{Vinyals_2015} & 0.663 & 0.423 & 0.277 & 0.183 & 0.237 & - & 0.855 & - \\
      m-RNN~\cite{mao2014deep} & 0.600 & 0.410 & 0.280 & 0.190 & 0.228 & - & 0.842 & - \\
      Soft-Attention~\cite{XuICML2015} & 0.707 & 0.492 & 0.344 & 0.243 & 0.239 & - & - & - \\
      Hard-Attention~\cite{XuICML2015} & 0.718 & 0.504 & 0.357 & 0.250 & 0.230 & - & - & - \\
      Semantic Attention~\cite{You_2016} & 0.709 & 0.537 & 0.402 & 0.304 & 0.243 & - & - & - \\
      Review Net~\cite{ReviewnetNIPS2016} & - & - & - & 0.290 & 0.237 & - & 0.886 & - \\
      LSTM-A5~\cite{yao2016boosting} & 0.730 & 0.565 & 0.429 & 0.325 & 0.251 & 0.538 & 0.986 & - \\
      \hline 
      \hline
      Encoder-Decoder & 0.718 & 0.547 & 0.412 & 0.311 & 0.251 & 0.530 & 0.961 & 0.179 \\
      Encoder-Decoder + Zoneout & 0.708 & 0.537 & 0.403 & 0.304 & 0.249 & 0.525 & 0.941 & 0.176 \\
      Encoder-Decoder + Scheduled Sampling & 0.718 & 0.548 & 0.414 & 0.315 & 0.252 & 0.531 & 0.975 & 0.180 \\
      Encoder-Decoder + ARNet & 0.730 & 0.562 & 0.425 & 0.321 & 0.252 & 0.535 & 0.988 & 0.182 \\
      \hline 
      \hline
      Attentive Encoder-Decoder & 0.727 & 0.557 & 0.421 & 0.318 & 0.259 & 0.537 & 0.996 & 0.185 \\
      Attentive Encoder-Decoder + Zoneout & 0.720 & 0.549 & 0.415 & 0.314 & 0.251 & 0.532 & 0.975 & 0.181 \\
      Attentive Encoder-Decoder + Scheduled Sampling & 0.731 & 0.563 & 0.426 & 0.322 & 0.256 & 0.538 & 1.006 & 0.187 \\
      Attentive Encoder-Decoder + ARNet & \textbf{0.740} & \textbf{0.576} & \textbf{0.440} & \textbf{0.335} & \textbf{0.261} & \textbf{0.546} & \textbf{1.034} & \textbf{0.190} \\
      \hline
    \end{tabular}
  \end{center}
  \vspace{-15pt}
\end{table*}

\section{Experimental Results}

\subsection{Image Captioning}
Image captioning is a task to generate a natural sentence to describe the visual content of one given image. In this paper, we use the most popular MSCOCO dataset~\cite{MSCOCO} to demonstrate the effectiveness of our proposed ARNet.

\subsubsection{Dataset}
The MSCOCO dataset contains 123,000 images with at least 5 captions for each image. We use the same data split as in~\cite{KarpathyPAMI2017} for performance comparisons, which reserves 5000 images for both validation and testing. We convert all captions into lowercase, remove non-alphanumeric characters, and tokenize the captions using white space. We keep the words that occur at least 5 times, resulting in a vocabulary size of 10,516. We truncate all the captions longer than 30 words. The beginning of each sentence is marked with a special \texttt{BOS} token, and the end with an \texttt{EOS} token.

\subsubsection{Implementation Details}
We take Inception-V4 model pre-trained on ImageNet as encoder. More specifically, we define the output of \emph{Average Pooling} layer in Inception-V4 network as the global information vector $g$, the output of the last \emph{Inception-C blocks} as local information vectors $\mathbf{s}$. In this case, $g$ is a vector with dimension $1536$, and $\mathbf{s}= \{ s_1, ..., s_{64} \}$ is a set containing 64 vectors with dimension $1536$. During the whole training stage, we do not finetune encoder. For decoder, LSTM unit with single layer is used. The dimensions of the hidden state and word embedding are set as 512. For training, the conventional encoder-decoder model is first trained until convergence by only considering the negative likelihood as defined in Eq.~\eqref{obj_lm}. Afterwards, the objective function defined in Eq.~\eqref{obj_whole} is used to train the proposed ARNet and finetune the encoder-decoder. During the first training stage, we use Adam~\cite{Adam} with an initial learning rate $5 \times 10^{-4}$. Then, we set the learning rate as $1 \times 10^{-4}$ to continue to train the model with ARNet. Early stopping is used to prevent overfitting. 
Beam search with size as 3 is utilized to generate the final caption for one given image.

\begin{figure*}[t]
 \centering
  \begin{tabular}{| c | m{6.2cm} | m{7.6cm} |}
    \hline
    Images & Generated Captions & Ground Truth Captions\\
    \hline 
    \hline
    \begin{minipage}{.12\textwidth}
      \includegraphics[width=\linewidth, height=0.6\linewidth]{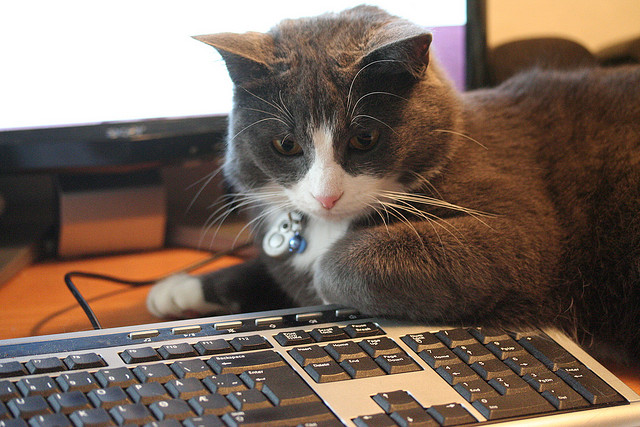}
    \end{minipage}
    &
	\begin{minipage}[t]{6.2cm}\scriptsize
        \textbf{Attentive Encode-Decoder}: \\
        a close up of a cat on a desk. \\ \\
        \textbf{Attentive Encode-Decoder-ARNet}:\\
        a cat sitting on a desk next to a \textcolor{red}{keyboard}.
    \end{minipage}
    &
    \begin{minipage}{7.6cm}\scriptsize
1. a grey cat peers at a computer keyboard. \\
2. a cat laying down by a keyboard. \\
3. a kitty playing with the keyboard on a laptop. \\
4. a large cat laying atop a computer keyboard. \\
5. a cat that is laying on a computer keyboard.
    \end{minipage}
    \\
    \hline
    \begin{minipage}{.12\textwidth}
      \includegraphics[width=\linewidth, height=0.6\linewidth]{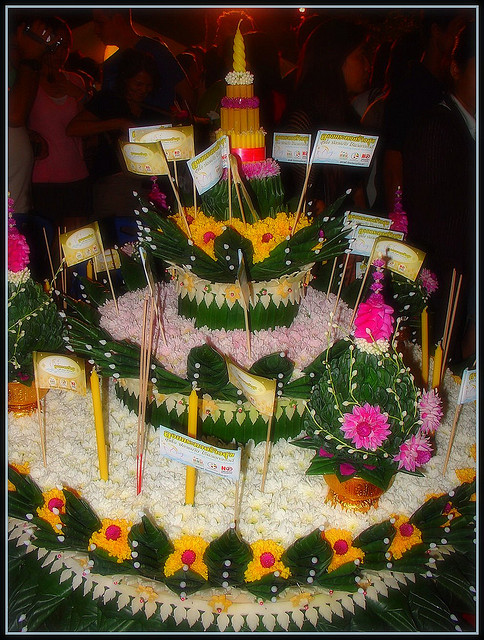}
    \end{minipage}
    &
	\begin{minipage}[t]{6.2cm}\scriptsize
        \textbf{Attentive Encode-Decoder}: \\
        a display of many different types of cake. \\ \\
        \textbf{Attentive Encode-Decoder-ARNet}:\\
        a cake decorated with many different types of \textcolor{red}{flowers}.
    \end{minipage}
    &
    \begin{minipage}{7.6cm}\scriptsize
1. a layered cake with many decorations on a table. \\
2. a large multi layered cake with candles sticking out of it. \\
3. a party decoration containing flowers, flags, and candles. \\
4. a cake decorated with flowers and flags on it. \\
5. a cake is decorated with flowers and flags.
    \end{minipage}
    \\
    \hline
    \begin{minipage}{.12\textwidth}
      \includegraphics[width=\linewidth, height=0.6\linewidth]{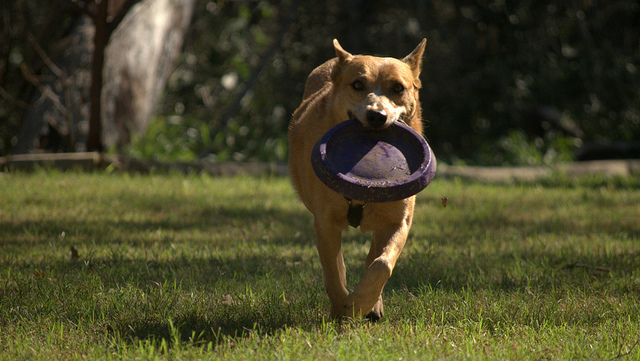}
    \end{minipage}
    &
	\begin{minipage}[t]{6.2cm}\scriptsize
        \textbf{Attentive Encode-Decoder}: \\
        a brown dog holding a blue frisbee in it's mouth. \\ \\
        \textbf{Attentive Encode-Decoder-ARNet}:\\
        a dog running in the \textcolor{red}{grass} with a frisbee in its mouth.
    \end{minipage}
    &
    \begin{minipage}{7.6cm}\scriptsize
1. a very cute brown dog with a disc in its mouth. \\
2. a dog running in the grass with a frisbee in his mouth. \\
3. a dog in a grassy field carrying a frisbee. \\
4. a brown dog walking across a green field with a frisbee in it's mouth. \\
5. a dog carrying a frisbee in its mouth running on a grass lawn.
    \end{minipage}
    \\
    \hline
    \begin{minipage}{.12\textwidth}
      \includegraphics[width=\linewidth, height=0.6\linewidth]{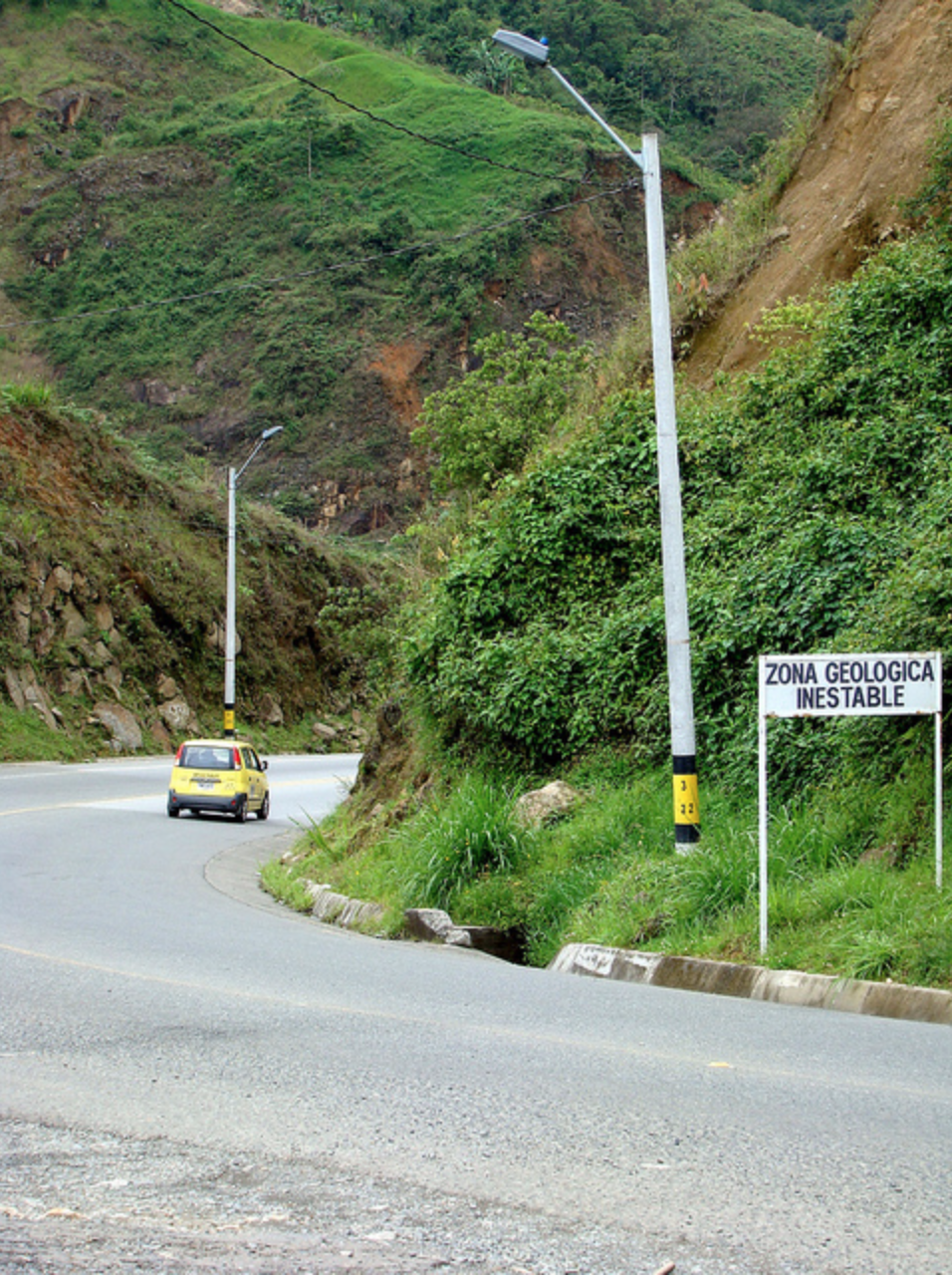}
    \end{minipage}
    &
	\begin{minipage}[t]{6.2cm}\scriptsize
        \textbf{Attentive Encode-Decoder}: \\
        a truck driving down a road next to a forest. \\ \\
        \textbf{Attentive Encode-Decoder-ARNet}:\\
        a car driving down a road next to a \textcolor{red}{lush green hillside}.
    \end{minipage}
    &
    \begin{minipage}{7.6cm}\scriptsize
1. a street scene of a road going through the mountains. \\
2. a road curving around hills has one car on it.  \\
3. a yellow car driving away on the road.  \\
4. a small yellow and black car driving around the bend of a road between.   \\
5. a small yellow car going around a turn and a sign.\\
    \end{minipage}
    \\
    \hline
  \end{tabular}
  \vspace{2pt}
  \caption{Example captions from the conventional model and our attentive encoder-decoder-ARNet model, along with their corresponding ground truth captions. It can be observed that ours can yield more detailed descriptions with meaningful words highlighted in boldface, such as ``\texttt{keyboard}'', ``\texttt{flowers}'', ``\texttt{grass}'',  and so on.}
  \label{caption-example}
  \vspace{-10pt}
\end{figure*}

\subsubsection{Evaluation and Comparison}
We use the MSCOCO evaluation toolkit\footnote{https://github.com/tylin/coco-caption} to compute BLEU~\cite{BleuScore}, METEOR \cite{Banerjee_METEOR_2005}, ROUGE-L~\cite{ROUGE}, and  CIDEr~\cite{CiderCVPR2015} scores to measure the quality of captions. Since SPICE~\cite{SpiceECCV2016} captures human judgments better than other automatic metrics, the resulting SPICE scores are also presented. Neural Image Caption~(NIC)~\cite{Vinyals_2015} and Soft Attention model~\cite{XuICML2015} are used as the encoder-decoder and attentive encoder-decoder for our proposed  ARNet. We also report the metric scores of models with scheduled sampling. Additionally, we also compare with m-RNN \cite{mao2014deep}, Semantic Attention \cite{You_2016}, Review Net \cite{ReviewnetNIPS2016}, and LSTM-A5~\cite{yao2016boosting}.
Table~\ref{table_1} shows the performance comparisons of different models. It can be observed  that ARNet can help improve the performance of both encoder-decoder and attentive encoder-decoder. Our proposed ARNet also outperforms scheduled sampling and zoneout, which can be also viewed as RNN regularizers. Moreover the attentive encoder-decoder with ARNet achieves the best performance. Therefore, the strategy forcing the current hidden state embedding more useful information from the past can more effectively regularize LSTM and thus improve the generated caption quality. 

Some qualitative results are shown in Fig.~\ref{caption-example}. It can be observed that the attentive encoder-decoder model with our proposed ARNet can generate more detailed and vivid descriptions for given images, such as the words ``\texttt{keyboard}'', ``\texttt{flowers}'', and so on.
\begin{figure}[!t]
  \centering
  \subfigure[]{\includegraphics[scale=0.40]{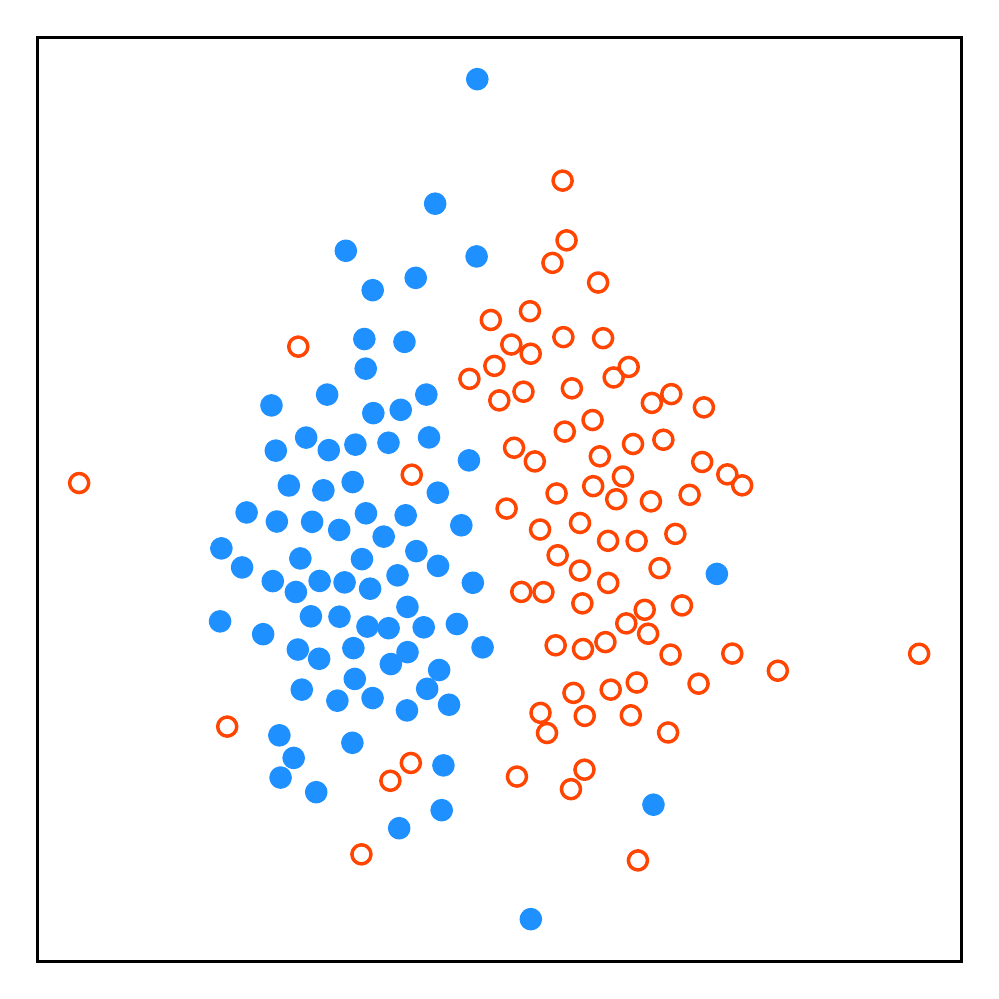}}
  \subfigure[]{\includegraphics[scale=0.40]{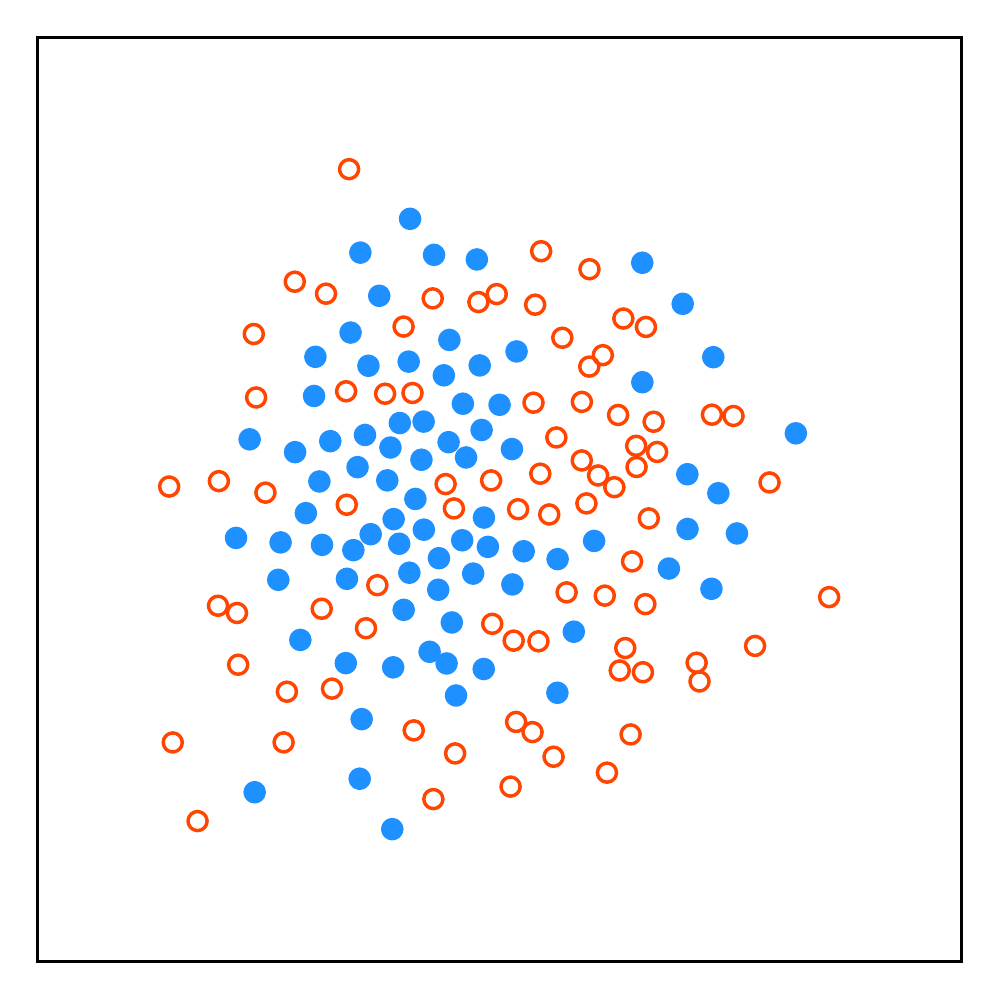}}
  \caption{Hidden states visualization of the attentive encoder-decoder model (a) and the attentive encoder-decoder-ARNet model (b). The filled circles in blue represent the hidden states generated in the training mode, while the open circles in red are obtained in the inference mode.}
  \vspace{-10pt}
  \label{fig3}
\end{figure}

\subsubsection{Discrepancy Analysis between Training and Inference}
Discrepancy between training and inference is a well known problem for RNN~\cite{ScheduledSampling,ProfessorForcing}. In the training stage, RNN is usually trained to maximize the likelihood of each token in the sequence given the current state and previous correct token from ground truth. At inference stage, the previous token is unknown and replaced by a token generated by the model itself. Hence, errors can be accumulated quickly along the generated sequence. To mitigate this problem, the distribution of sequences of training and inference state should be non-distinguishable. Here, to study this problem, we consider the distributions of last hidden states of sequences as in \cite{ProfessorForcing}, since they encode the necessary information about the whole sequence.

We extract the hidden state of the LSTM unit which emits the \texttt{EOS} token or reaches the maximum time step. We visualize one batch with T-SNEs~\cite{maaten_t-SNE} both for training and inference, where the batch size is $80$. Fig. \ref{fig3} shows the T-SNE visualization of hidden states for attentive encoder-decoder model and attentive encoder-decoder-ARNet model. We can see that our ARNet can significantly reduce the discrepancy between training and inference. We believe that it is one of the reasons why models with ARNet perform better than the counterparts.

For further evaluating the discrepancy quantitatively, a appropriate metric is needed. Since the hidden states are from different models lying in different spaces, computing the Euclidean distance between them is not reasonable. In this paper, we thereby consider cosine distance between hidden states, which is defined as follows:
\begin{align}
	d(h_1, h_2) = 1 - \frac{h_1^T h_2}{\|h_1\|\|h_2\|}.
\end{align}
The cosine distance considers the angle between $h_1$ and $h_2$, which will not be affected by the norm of $h_1$ and $h_2$. 

\begin{table}[!t]
  \caption{Discrepancy between training and inference modes on image captioning task measured by the mean centroid and point-wise distances defined in Eqs. (\ref{eq:mc}) and (\ref{eq:pw}). Smaller distance values indicate better performances.}\label{table_2}
  \vspace{-5pt}
  \begin{center}
    \small
    \begin{tabular}{|l|c|c|}
      \hline
      Model Name & ${d_{mc}}$ & ${d_{pw}}$ \\
      \hline 
      \hline
      Encoder-Decoder & 0.747 & 0.719 \\
      Encoder-Decoder + ARNet & 0.514 & 0.561 \\
      \hline 
      \hline
      Attentive Encoder-Decoder & 0.773 & 0.760 \\
      Attentive Encoder-Decoder + ARNet & \textbf{0.491} & \textbf{0.595} \\
      \hline
    \end{tabular}
  \end{center}
  \vspace{-10pt}
  \label{image_caption_distance}
\end{table}

\begin{table*}[!t]
  \caption{Performance comparison on the testing split of the HabeasCorpus dataset. The best results among all models are marked with boldface.}
  \vspace{-5pt}
  \begin{center}
  \small
  \tabcolsep=0.275cm
  \begin{tabular}{|l|c|c|c|c|c|c|}
    \hline
    Model Name & BLEU-1 & BLEU-2 & BLEU-3 & BLEU-4 & METEOR & ROUGE-L\\
    \hline 
    \hline
    Review Net~\cite{ReviewnetNIPS2016} & 0.192 & 0.105 & 0.074 & 0.057 & 0.085 & 0.200 \\
    \hline 
    \hline
    Encoder-Decoder & 0.183 & 0.093 & 0.063 & 0.047 & 0.080 & 0.188 \\
    Encoder-Decoder + Zoneout & 0.182 & 0.080 & 0.063 & 0.047 & 0.080 & 0.181 \\
    Encoder-Decoder + Scheduled Sampling & 0.186 & 0.098 & 0.067 & 0.051 & 0.082 & 0.194 \\
    Encoder-Decoder + ARNet & 0.196 & 0.107 & 0.075 & 0.058 & 0.089 & 0.213 \\
    \hline 
    \hline
    Attentive Encoder-Decoder & 0.228 & 0.140 & 0.106 & 0.088 & 0.105 & 0.256 \\
    Attentive Encoder-Decoder + Zoneout & 0.227 & 0.140 & 0.105 & 0.086 & 0.090 & 0.220 \\
    Attentive Encoder-Decoder + Scheduled Sampling & 0.229 & 0.142 & 0.108 & 0.089 & 0.107 & 0.270 \\
    Attentive Encoder-Decoder + ARNet & \textbf{0.255} & \textbf{0.173} & \textbf{0.139} & \textbf{0.120} & \textbf{0.123} & \textbf{0.289} \\
    \hline
  \end{tabular}
  \end{center}
  \vspace{-15pt}
  \label{habeascorpus}
\end{table*}

Based on cosine distance, we define two different distance metrics to measure these different models.
More specifically, let $\mathbf{U} = \{ u_{\text{I}_1}, ..., u_{\text{I}_B} \}$, $\mathbf{V} = \{ v_{\text{I}_1}, ..., v_{\text{I}_B} \}$ be the last hidden states of decoder that we get from training and inference modes given input images $\text{I}_1, \text{I}_2, \cdots, \text{I}_B$, respectively. The first distance metric is the mean centroid distance $d_{\text{mc}}$:
\begin{equation}
\label{eq:mc}
d_{\text{mc}}\left( \mathbf{U}, \mathbf{V} \right) = d\left(\frac{1}{B}\sum_{i}^{B}{u_{\text{I}_i}}, \frac{1}{B}\sum_{j}^{B}{v_{\text{I}_i}} \right).
\end{equation}
The second distance metric $d_{\text{pw}}$ is the point-wise distance between the hidden states of the same input but from training and inference respectively. And $d_{\text{pw}}$ can be computed according to:
\begin{equation}
\label{eq:pw}
  d_{\text{pw}} \left( \mathbf{U}, \mathbf{V} \right) = \frac{1}{B} \sum_{i=1}^{B}{ d\left(u_{\text{I}_i}, v_{\text{I}_i} \right)}.
\end{equation}
$d_{\text{pw}}$ only measures the difference between the ground-truth and sequence generated from the same image. 
By considering the two distances, a more accurate study of the discrepancy between training and inference is conducted.

Table~\ref{table_2} shows the discrepancies between training and inference of different models, measured by $d_{\text{mc}}$ and $d_{\text{pw}}$. It can be clearly observed that our ARNet yields smaller differences between the representations of ground-truth and sequence generated for the same image. Thus ARNet can significantly reduce the discrepancies of the encoder-decoder and attentive encoder-decoder models. As such, the generated sequences  are more semantically similar to the ground-truth.

\subsubsection{Effect of $\lambda$}
\begin{figure}[t]
  \centering
  \includegraphics[width=1\linewidth]{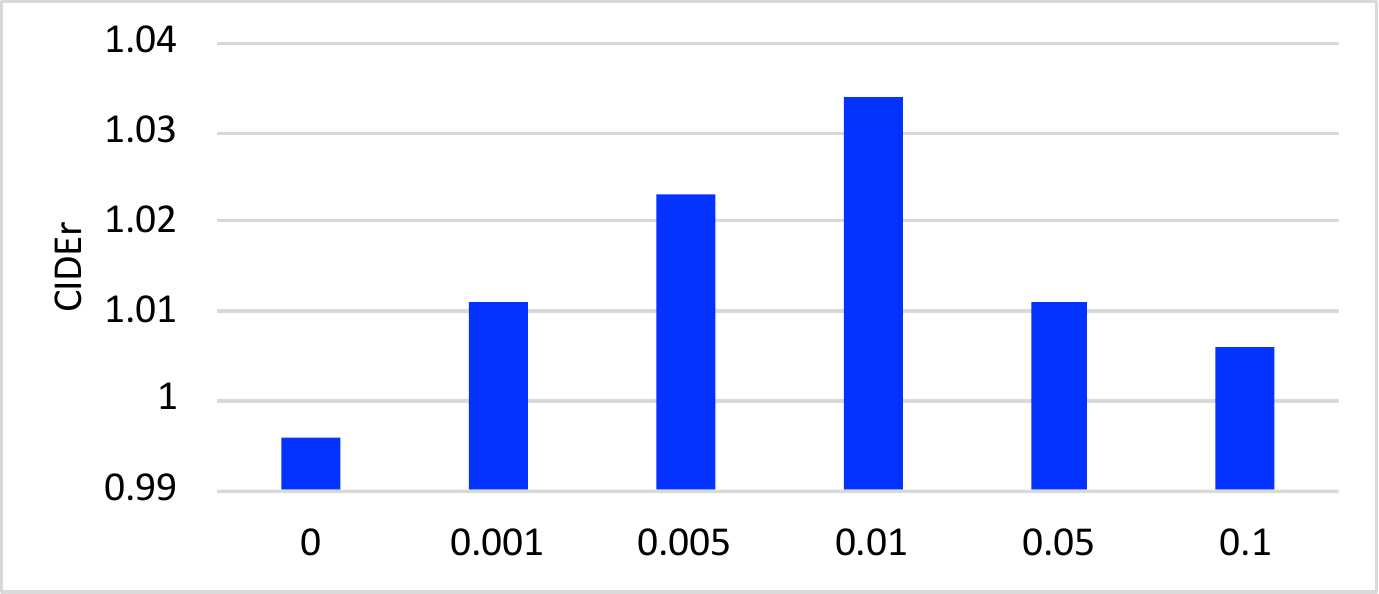}
  \caption{The CIDEr scores with different $\lambda$ weights as in Eq. (\ref{obj_whole}), ranging in \{0, 0.001, 0.005, 0.01, 0.05, 0.1\}. The first bin with $\lambda=0$ denotes the vanilla attentive encoder-decoder model.}
  \vspace{-10pt}
  \label{lambda_effect}
\end{figure}

The parameter $\lambda$ balances the contributions from the encoder-decoder and ARNet. If $\lambda$ is set as 0, our model downgrades as the conventional encoder-decoder model. Different $\lambda$ values are evaluated. Fig. \ref{lambda_effect} shows CIDEr scores of attentive encoder-decoder-ARNet models with different $\lambda$.
Our model with these positive $\lambda$ values always performs better than the conventional encoder-coder model, which proves that ARNet with the regularization on the transition dynamics is effective to improve the image captioning performance. If $\lambda$ is too large, the performance will decrease, since the model focuses too much on the reconstruction part and ignores the supervision signal from ground truth. To achieve better performance, appropriate $\lambda$ needs to be carefully selected on the validation set. In this paper, $\lambda$ is experimentally chosen as 0.01 for the image captioning task.

\subsection{Code Captioning}

\subsubsection{Dataset}
For the code captioning task, We utilize the HabeasCorpus~\cite{MovshovitzACL2013} dataset which is collected from nine open source JAVA projects and contains $6,734$ source code files. Following the public split \cite{ReviewnetNIPS2016}, the training, validation and testing datasets, containing $5,370$, $702$ and $662$ files, respectively, are used for our experiments. Each source code sequence is associated with a comment sentence which summarizes the intention of the file. We transform the code comment sentences into lowercase, tokenize them with white space, resulting in a vocabulary with size $12,860$. We truncate all the code sequences and comment sentences such that they have 300 tokens at most. 
BLEU, METEOR, and ROUGE-L are also used to measure the relevance with respect to the reference sentences.

\subsubsection{Implementation Details}
We realize our ARNet on both the plain and attentive encoder-decoder frameworks. The encoder and decoder network are both single layer LSTM with hidden unit size 256. The word embedding size is 512. We pre-train the model without ARNet with learning rate $1 \times 10^{-3}$. Then we train the whole model with learning rate  $5 \times 10^{-4}$. The batch size is set as 16. And the training procedure is terminated with early stopping strategy when BLEU-4 score reaches the maximum value on the validation set.

\vspace{-0.1cm}
\subsubsection{Evaluation and Comparison}
Table~\ref{habeascorpus} summarizes the results on the testing set of HabeasCorpus dataset. We implement all the models and report the performances under the same settings. Our attentive ARNet and non-attentive ARNet achieve $36.36\%$ and $23.40\%$ relative improvements  on BLEU-4 metric over baseline model, respectively. Again, our method significantly outperforms scheduled sampling and zoneout.  Moreover, comparing with image captioning, the improvements brought by ARNet is even more significant. The main reason may due to the time step length of the decoder. Our proposed ARNet make connections between neighboring hidden states by the reconstruction strategy, which effectively regularize the transition dynamics. Therefore, with time step increasing, ARNet can make more effective gradient information flow, compared to plain decoder.

\vspace{-0.1cm}
\subsubsection{Discrepancy Analysis}
To study the discrepancy between training and inference on this task, we also compute the distances measured by $d_{\text{mc}}$ and $d_{\text{pw}}$. The results of different models are shown in Table~\ref{code_caption_distance}. Similarly, we can observe that that our ARNet can help mitigate the discrepancy between training and inference, thus making the inference more robust and improving the quality of generated code captions.

\begin{table}[!t]
  \caption{Discrepancy between training and inference modes on code captioning task measured by the mean centroid and point-wise distances defined in Eqs. (\ref{eq:mc}) and (\ref{eq:pw}). Smaller distance values indicate better performance.}
  \vspace{-5pt}
  \begin{center}
    \begin{tabular}{|l|c|c|}
      \hline
      Model Name & ${d_{mc}}$ & ${d_{pw}}$ \\
      \hline \hline
      Encoder-Decoder & 0.643 & 0.722 \\
      Encoder-Decoder + ARNet & 0.641 & 0.699 \\
      \hline \hline
      Attentive Encoder-Decoder & 0.594 & 0.712 \\
      Attentive Encoder-Decoder + ARNet & \textbf{0.322} & \textbf{0.465} \\
      \hline
    \end{tabular}
  \end{center}
  \vspace{-18pt}
  \label{code_caption_distance}
\end{table}

\subsection{Permuted Sequential MNIST}
In this section, in order to further examine the regularizing ability of our proposed ARNet on modeling long term dependencies, a new task, namely permuted sequential MNIST~\cite{le2015simple,ZoneoutICLR2017}, is considered.
Sequential MNIST is first proposed~\cite{le2015simple} to classify MNIST digits, when the 784 pixels are presented sequentially to the recurrent net. Permuted sequential MNIST is an even more challenging problem, with the pixels presented in a (fixed) random order. 


The permuted pixel sequence is encoded by one single LSTM layer with hidden size of 128. As introduced in Sec.~\ref{sec:architecture}, ARNet is realized by another LSTM, coupling with the encoder, to further regularize the LSTM transition dynamics. In this paper, the hidden size in ARNet is also 128. The training is performed in two stages. We first make pre-training on the encoder LSTM. Afterwards, the two LSTMs of encoder and ARNet are jointly trained. Adam~\cite{Adam} with learning rate $1\times10^{-3}$ and $5\times10^{-4}$ are used for the two stages, respectively. The batch size is set as 64.

Besides the unregularized LSTM, we also compare with the other two regularziers, specifically the recurrent dropout~\cite{dropout} and zoneout~\cite{ZoneoutICLR2017}. The performance comparisons are shown in Table~\ref{pMNIST}, where the test accuracies of all models are reported. First, the permuted sequential MNIST is much more challenging, and LSTM can only achieve 91.4\% accuracy. But by incorporating different regularizers, the test accuracies can be significantly improved.  Moreover, with coupling ARNet with the unregularized LSTM, we outperforms the recurrent dropout and zoneout. The encouraging results on permuted sequential MNIST task shows our ARNet can model long term dependencies more effectively in the data. 

\begin{table}[!t]
  \caption{Performance comparisons on permuted sequential MNIST task. Our proposed ARNet outperforms recurrent dropout and zoneout.}
  \vspace{-5pt}
  \begin{center}
    \begin{tabular}{|l|c|}
      \hline
      Model Name & Test Accuracy \\
      \hline \hline
      LSTM + recurrent dropout  & 0.925 \\
      LSTM + zoneout & 0.931 \\
      \hline \hline
      Unregularized LSTM & 0.914 \\
      LSTM + ARNet & \textbf{0.933} \\
      \hline
    \end{tabular}
  \end{center}
  \vspace{-18pt}
  \label{pMNIST}
\end{table}

\section{Conclusion}
In this paper, aiming at regularizing the transition dynamics and mitigating the discrepancy of RNN for sequence prediction, a novel auto-reconstructor network (ARNet) was proposed. ARNet, coupling with the conventional encoder-decoder framework, reconstructs the past hidden state with the current one, thus encouraging the present hidden state to embed more information from the previous one. As such, ARNet can  improve the performance of various caption generation tasks. The extensive experimental results on image captioning, source code captioning, and permuted sequential MNIST tasks demonstrate the superiority of our proposed ARNet.

\vspace{-5pt}
\section*{Acknowledgement}
\vspace{-5pt}
This work was partially supported by the National Key Research and Development Program of China (Project No.~2017YFB1302400) and the National Natural Science Foundation of China (Project No. 41571436).

{\small
\bibliographystyle{ieee}
\bibliography{egbib}
}

\end{document}